\documentclass[]{spie}  

\usepackage{amsmath,amsfonts,amssymb}
\usepackage{tabularx}
\usepackage{graphicx}
\usepackage{booktabs}
\usepackage{xcolor,colortbl}
\usepackage{float}
\usepackage{afterpage}
\definecolor{deepgreen}{RGB}{10,90,40} 
\definecolor{lessred}{RGB}{210,100,70}  
\definecolor{lessblue}{RGB}{100,100,230} 
\definecolor{lightgray}{gray}{0.9} 
\usepackage[colorlinks=true, allcolors=blue]{hyperref}

\title{Weighted Circle Fusion: Ensembling Circle Representation from Different Object Detection Results}

\author[a]{Jialin Yue}
\author[a]{Tianyuan Yao}
\author[a]{Ruining Deng}
\author[a]{Quan Liu}
\author[a]{Juming Xiong}
\author[a]{Junlin Guo}
\author[b]{Haichun Yang}
\author[a]{Yuankai Huo}

\affil[a]{Vanderbilt University, Nashville, TN, USA}

\affil[b]{Vanderbilt University Medical Center, Nashville, TN, USA}

\authorinfo{Corresponding author: Yuankai Huo: E-mail: yuankai.huo@vanderbilt.edu}

\begin{document} 
\maketitle

\begin{abstract}
Recently, the use of circle representation has emerged as a method to improve the identification of spherical objects (such as glomeruli, cells, and nuclei) in medical imaging studies. In traditional bounding box-based object detection, combining results from multiple models improves accuracy, especially when real-time processing isn't crucial. Unfortunately, this widely adopted strategy is not readily available for combining circle representations. In this paper, we propose Weighted Circle Fusion (WCF), a simple approach for merging predictions from various circle detection models. Our method leverages confidence scores associated with each proposed bounding circle to generate averaged circles. We evaluate our method on a proprietary dataset for glomerular detection in whole slide imaging (WSI) and find a performance gain of 5\% compared to existing ensemble methods. Additionally, we assess the efficiency of two annotation methods—fully manual annotation and a human-in-the-loop (HITL) approach—in labeling 200,000 glomeruli. The HITL approach, which integrates machine learning detection with human verification, demonstrated remarkable improvements in annotation efficiency. The Weighted Circle Fusion technique not only enhances object detection precision but also notably reduces false detections, presenting a promising direction for future research and application in pathological image analysis. The source code has been made publicly available at https://github.com/hrlblab/WeightedCircleFusion
\end{abstract}


\section{INTRODUCTION}
\label{sec:intro}  

\begin{figure*}[t]
\begin{center}
\includegraphics[width=1\textwidth]{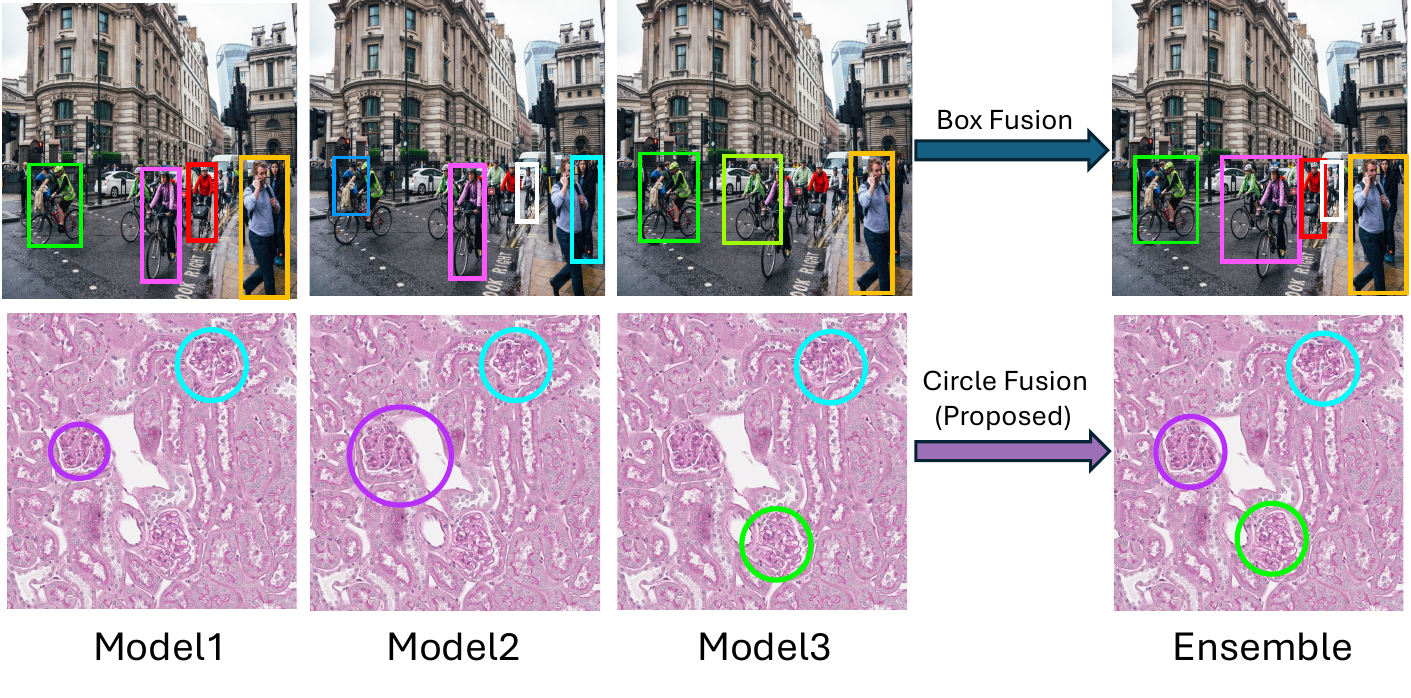}
\end{center}
\caption{Comparison of Box Fusion and Circle Fusion Methods for Object Detection. This figure delineates the differences between the ensemble results of box representation and circle representation. Box fusion alters the dimensions of the box, thereby changing its shape, while circle fusion only modifies the radius of the circle, preserving its shape. For the detection of medical ball-shaped objects, circle representation can achieve better performance.} 
\label{fig1:Problem}
\end{figure*}

Object detection plays an essential role in medical imaging~\cite{jaeger2020retina}, offering a wide range of applications that are enhanced by machine learning technologies. Traditional object detection models, such as Faster R-CNN~\cite{girshick2015fast}, YOLO~\cite{redmon2016you}, and SSD~\cite{liu2016ssd}, have been widely adopted across various domains for their efficiency and accuracy~\cite{jiang2017face}. In medical object detection tasks, detecting glomeruli is essential for effective diagnosis and quantitative assessments in renal pathology. For these tasks, CircleNet~\cite{yang2020circlenet} stands out in the medical field for its unique approach to detection tasks. Unlike conventional detection networks that rely on bounding boxes, CircleNet offers a rotation-consistent circle representation with fewer parameters for ball-shaped objects~\cite{xiong2024circle}, such as glomeruli in kidney pathology (Fig.~\ref{fig1:Problem}). Despite CircleNet's advantages, relying on a single CircleNet-trained model for detection tasks presents considerable challenges, including missed and false detections~\cite{natu2021privacy}.

To enhance the robustness of object detection, ensemble learning algorithms, such as Non-Maximum Suppression (NMS)~\cite{1699659}, Soft-NMS~\cite{bodla2017soft}, and Weighted Box Fusion (WBF)~\cite{solovyev2021weighted}, have been proposed to fuse the detection results from multiple models (Fig.~\ref{fig1:Problem}). NMS and Soft-NMS work by eliminating lower confidence detections based on an Intersection Over Union (IOU) threshold~\cite{rezatofighi2019generalized}, with Soft-NMS adjusting detection scores rather than removing detections outright. WBF further refines this approach by merging overlapping detections, allowing those with higher confidence scores to improve the merged result. Unfortunately, such methods were optimized for traditional bounding box based representation for natural images.

In this paper, we propose a simple ensemble method, called Weighted Circle Fusion (WCF), designed specifically for circle representation in medical imaging detections. This method merges overlapping detections, with the fusion result's position decided by the confidence of the contributing detections. Importantly, it calculates the number of overlapped circles merged for each object, while computing the average score for false positive elimination. In experiments, we assessed the detection results of glomeruli on whole slide images (WSIs) using five-fold cross-validation. Additionally, to validate the method's consistency across rotations, we tested it on images rotated by 90 degrees. The results demonstrate the method's decent rotation consistency. To summarize, the contribution of this paper is threefold:

 
$\bullet$ The WCF method, combined with a dual thresholds strategy, enhances precision and reliability by fusing detection results from circle representation and eliminating false positives based on confidence scores and overlap across hard decisions.

$\bullet$ Our method achieved a substantial performance gain ($>$ 5\% ) compared to the average results of individual models.

$\bullet$ Utilizing a human-in-the-loop (HITL) approach to test the time required to annotate 10 WSIs, showed that it saves  68.59\% of total annotation time compared to complete manual annotation.


\begin{figure*}[h]
\begin{center}
\includegraphics[width=1\textwidth]{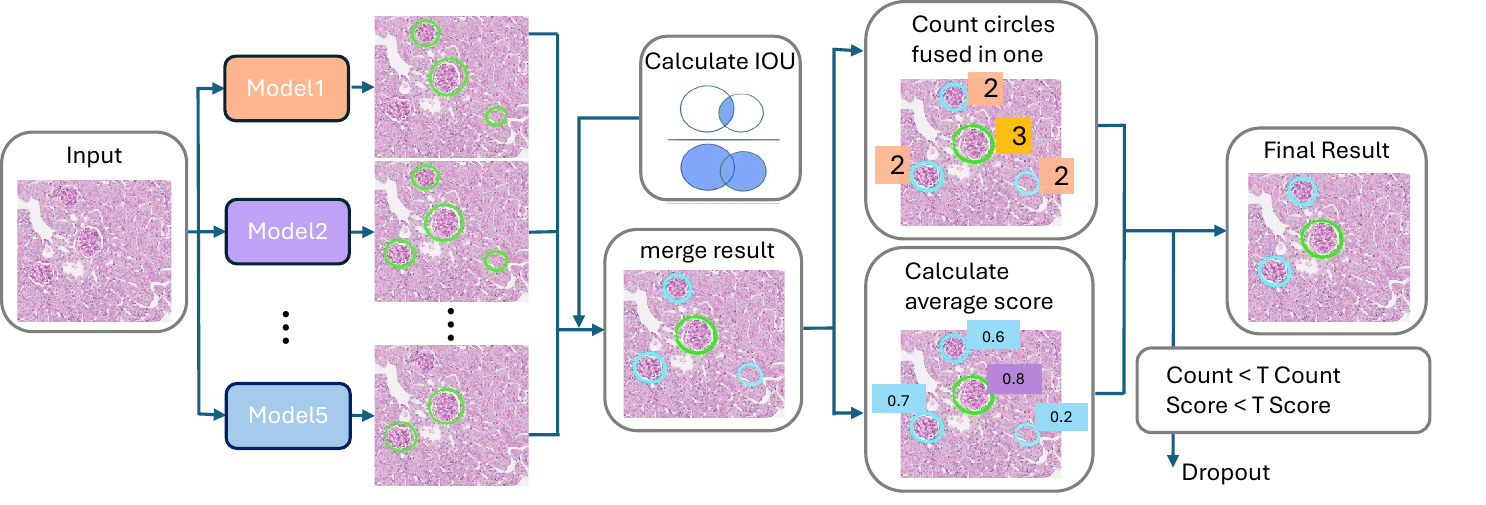}
\end{center}
\caption{The workflow of the proposed Weighted Circle Fusion (WCF) method. This figure delineates the specific steps involved in our method. The core of the method lies in counting the number of fused circles and calculating their average score, which is then used to eliminate potential erroneous detections. }
\label{fig2:method}
\end{figure*}
\vspace{-1em} 

\section{Methods}

In this section, we introduce an innovative method for fusing predictions: Weighted Circle Fusion (Fig.~\ref{fig2:method}). This technique is designed to enhance the accuracy of object detection, particularly focusing on circular objects commonly encountered in medical imaging, such as cells, glomeruli, or other spherically shaped features. Our approach involves pairwise fusion of the detection results from five models, where the results from the first model are fused with the second, then the combined results are fused with the third model, and so on until the fifth model is included.

The WCF process begins with aggregating predictions from multiple models, resulting in several sets of detection outcomes. Initially, the detection results from the first model are stored in a list, referred to as $R$. Subsequent detections from other models are compared against the entries in list $R$ based on their cIOU~\cite{yang2020circlenet}.The definition of cIOU can be found in the corresponding reference. If the cIOU between any two detections exceeds a predetermined threshold, indicating an enhanced agreement between models on the presence and location of an object, these detections are considered for fusion.

Upon fusion of the two results, it is necessary to recalculate the coordinates and confidence score of the new, combined result. Given that our detection results are represented as circles, we utilize the circles' center coordinates and radii for computation. Suppose the center coordinates and radius of a result from the first set are 
($x_1$,$y_1$) and $r_1$ with a confidence score $s_1$; and similarly, ($x_2$,$y_2$) and $r_2$ with score $s_2$ for a result from the second set. The formulas for calculating the weighted average coordinates and radius are as follows:

For center coordinates:

\begin{equation}
x_{\text{fuse}} = \frac{x_1 \cdot s_1 + x_2 \cdot s_2}{s_1 + s_2}
\label{eq:center_x}
\end{equation}

\begin{equation}
y_{\text{fuse}} = \frac{y_1 \cdot s_1 + y_2 \cdot s_2}{s_1 + s_2}
\label{eq:center_y}
\end{equation}

For radius:
\begin{equation}
r_{\text{fuse}} = \frac{r_1 \cdot s_1 + r_2 \cdot s_2}{s_1 + s_2}
\label{eq:radius}
\end{equation}

After calculating the fused coordinates, we compute the average of the scores of the merged results and keep track of how many detections have been merged to form this new result.

If a result from the second set cannot fuse with any result in list $R$, it is directly added to $R$. This process is repeated for each set of predictions until all m sets have been processed.

Upon completing the fusion of all model predictions, the confidence score $S$ for the fused result is calculated as follows:
\begin{equation}
S = \frac{\sum_{m=1}^{M} S_m}{M}
\label{eq:confidence_score}
\end{equation}
 
where $S_m$ is the confidence score of each individual model's prediction.

Additionally, we apply a ``count score'' $C$ to quantify how many model predictions have been fused into a single detection. The max value of $C$ depends on how many models we use in our ensemble method.

To further refine the detection outcomes, we introduced two thresholds: ``T count'' for the count value $C$ and ``T score'' for the average score of each result. Specifically, if both the count value and average score are below their respective thresholds, the detection result will be discarded. For the experiments in this paper, "T count" is set to 2 and "T score" is set to 0.9. This strategic approach enhances the precision of detection, making WCF particularly effective for instances where erroneous detections are common.

\section{Experiments}
 \subsection{Data}

For our training dataset, we utilized an in-house dataset. This included 15,190 patches from whole slide images derived from renal biopsies. Additionally, we incorporated 9,260 patches from PAS-stained WSIs of murine kidneys. This dataset was divided into training, validation, and testing sets with a ratio of 7:1:2 for each of the five models.

For the training dataset for the plus version models, an additional 100,000 glomeruli were added to the basic training dataset used to train the base version of the model. These additional glomeruli were sourced from 170 WSI from our in-house dataset. The 100,000 glomeruli were divided into five groups of 40,000 glomeruli, with each group added to a different model. Each group of 40,000 glomeruli had a 20,000 overlap with the others. All patches in our training dataset were either cropped or resized to dimensions of 512 × 512 pixels. Each patch contained at least one glomerulus. 

To evaluate the efficiency of different annotation methods for 200,000 glomeruli, we compared fully manual annotation with a human-in-the-loop (HITL) approach. The manual method involved human experts marking each glomerulus, whereas the HITL method integrated machine learning detection with human verification and correction. This comparison was conducted to assess the time efficiency and effectiveness of incorporating machine learning into the annotation process.

For the testing dataset, we included 15 PAS-stained WSIs, encompassing 2051 mouse glomeruli.
\subsection{Experiment Setting}
The models were trained on the CircleNet architecture with a dla-34 backbone, using slightly varied datasets to enhance learning diversity and robustness. Training spanned 30 epochs for each model, and outputs were refined using the Non-Maximum Suppression algorithm.

We evaluated the efficiency of two annotation methods for 200,000 glomeruli in our KidneyPath dataset: fully manual annotation and a human-in-the-loop (HITL) approach. The manual method involved human experts marking each glomerulus, while the HITL method combined machine learning detection with human verification and correction. This comparison aimed to assess the time efficiency of integrating machine learning into the annotation process.
\subsubsection{Fusion Method Comparison Experiments}
In this part of the experiment, we compared three ensemble methods: NMS, Soft-NMS, and WCF, as well as the results from five models and their plus version.
Each model was enhanced by the addition of 40,000 glomeruli training data, leading to improved performance. These 40,000 glomeruli were derived from an additional collection of 100,000 glomeruli, with a 20,000 overlap between each model.

Our WCF method was configured with specific parameters: a circle Intersection Over Union (cIOU) threshold of 0.5. For the experiments in this paper, "T count" is set to 2 and "T score" is set to 0.9. Initially, the WCF algorithm was applied to the outputs refined by the NMS algorithm to combine the strengths of individual detections into a single, more accurate result. The effectiveness of the WCF-fused results was meticulously evaluated and compared against the performance of individual models, traditional NMS, and Soft-NMS, with cIOU thresholds set at 0.5 and 0.3, respectively.
\subsubsection{Rotational Consistency Experiments}
In this part, we delved into assessing the rotational consistency of their fusion method. This was achieved by extracting patches from Whole Slide Images and rotating them by 90 degrees prior to the detection process. The results from these rotated patches were then subjected to the same fusion process.

\subsubsection{Evaluation}
The models were evaluated based on the mean average precision (mAP) at IoU values of 0.5 and 0.75. Additionally, mAP was computed across a spectrum of IoU thresholds, thereby conducting a comprehensive assessment. This metric was calculated over a range of IoU thresholds, from 0.5 to 0.95 in steps of 0.05, at each step averaging the precision. Alongside precision, the average recall across these IoU thresholds was also measured, providing a rounded evaluation of model performance.

The IoU metric, a ratio reflecting the overlap between two objects versus their combined area, is traditionally calculated for bounding box representations. However, given that this study's predictions utilize circle representations, we adopted the circle IoU (cIoU)~\cite{nguyen2021circle} metric as our evaluation standard. The cIoU offers a more fitting measure for our circular detection outputs, aligning with the unique geometry of the objects being detected.

\begin{figure*}[h]
\begin{center}
\includegraphics[width=0.9\textwidth]{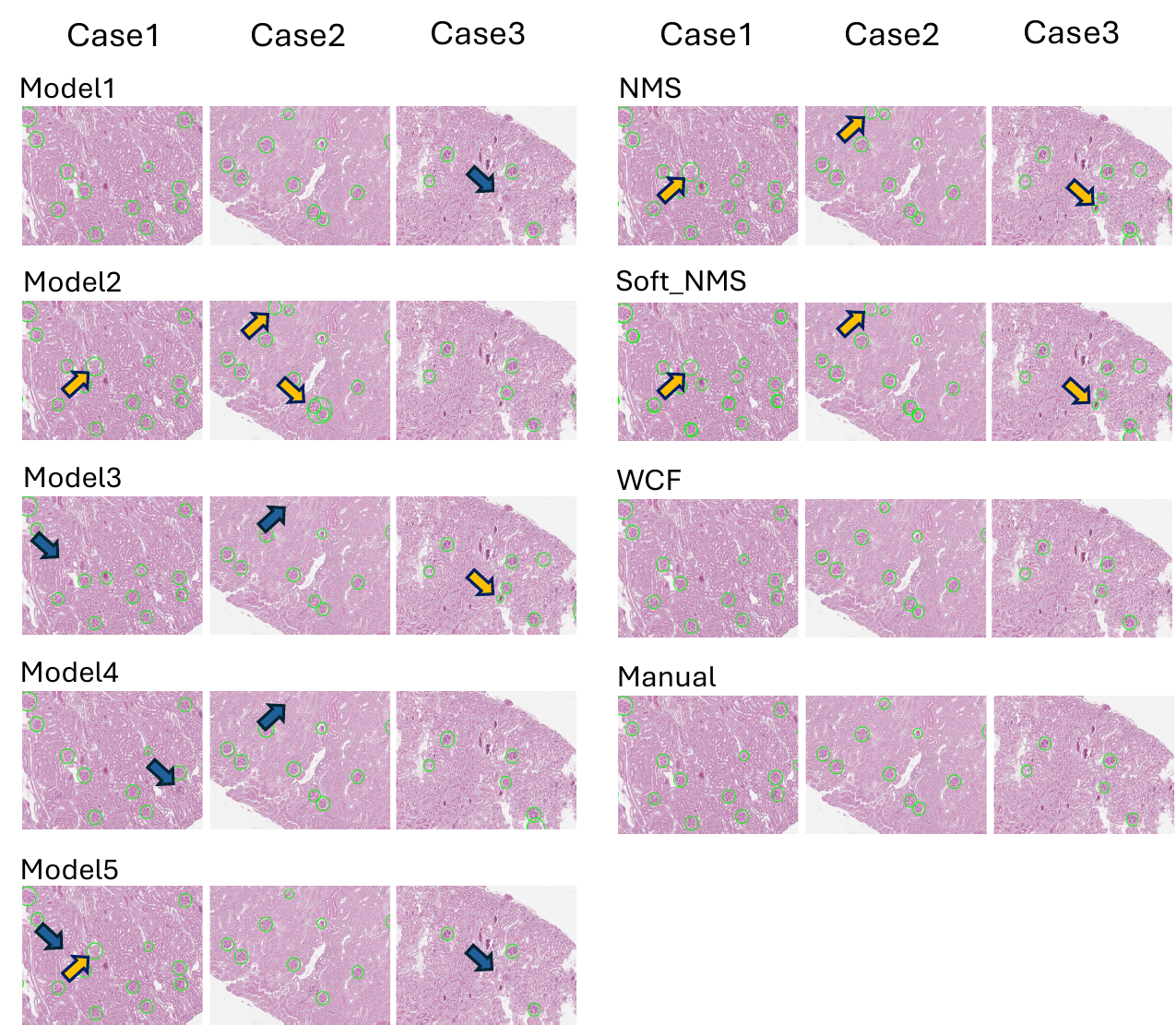}
\end{center}
\caption{Result Visualization. This figure presents the detection outcomes of glomeruli on WSIs using our method. The yellow arrows highlight false negatives identified by other models or methods, while the blue arrows indicate false positives. It is evident that traditional fusion methods such as NMS and soft-NMS tend to merge more erroneous predictions. In contrast, the WCF method achieves superior fusion results, with fewer incorrect predictions and the inclusion of detections that individual models failed to identify, demonstrating its effectiveness in enhancing detection accuracy. } 
\label{fig3:Annotation}
\end{figure*}

\section{Results}
\subsection{Performance on glomerular detection}
Fig.~\ref{fig3:Annotation} and Table~\ref{table:fusionresult}showcase the performance of our fusion method, which integrates the outputs from five models and their enhanced version on murine glomerular WSIs. Averaged results are calculated from five original models and five enhanced models with 40000 additional global features, providing a comprehensive comparison across different fusion methods. The results demonstrate that our approach achieves remarkably higher mAP values and average recall rates. The enhanced models exhibit better average recall and average precision compared to the original models. Notably, the mAP obtained through our method surpasses that of any individual model included in the study. Although the average recall of our method is slightly lower compared to other fusion methods, it remains competitively high and exceeds the average recall of the five original models.

\begin{table}[h]
\centering
\resizebox{0.9\textwidth}{!}{
\begin{tabular}{lcccc}
\toprule
Model & mAP(0.5:0.95) & mAP(@0.5IOU) & mAP(@0.75IOU) & Average Recall(0.5:0.95) \\ \hline\hline
CircleNet~\cite{yang2020circlenet}   & 0.594   & 0.784   & 0.676   & 0.605 \\ 

CircleNet+   &\textbf{ 0.764} &\textbf{ 0.899} &\textbf{ 0.825} &\textbf{ 0.738 }\\
\hline
NMS~\cite{1699659}  & 0.463  & 0.566  & 0.516  & 0.745  \\ 
NMS+  & 0.644  & 0.749  & 0.696  & \textbf{\textcolor{lessred}{0.834}}  \\ 
Soft-NMS~\cite{bodla2017soft}  & 0.319  & 0.402  & 0.357  & 0.722 \\
Soft-NMS+ & 0.419  & 0.513  & 0.452  & \textbf{\textcolor{lessblue}{0.793}} \\
WCF(Ours)  & \textbf{\textcolor{lessblue}{0.707}}  & \textbf{\textcolor{lessblue}{0.907}}  & \textbf{\textcolor{lessblue}{0.810}}  & 0.629  \\ 
WCF+(Ours)  & \textbf{\textcolor{lessred}{0.829}}  & \textbf{\textcolor{lessred}{0.955}}  & \textbf{\textcolor{lessred}{0.905}}  & 0.782  \\ 
\bottomrule
\end{tabular}
}
\caption{The table shows the averaged performance metrics of five original models ("Models in fold") and their enhanced versions with 40,000 additional global features ("Models+ in fold"). Metrics include mean average precision (mAP) at various IoU thresholds and average recall, evaluated using NMS, soft-NMS, and WCF fusion methods. Results highlight the superior performance of the WCF method across models.}

\label{table:fusionresult}
\end{table}

\begin{table}[h]
\centering
\begin{tabular}{lcccc}

\toprule
Model & mAP(0.5:0.95) & mAP(@0.5IOU) & mAP(@0.75IOU) & Average Recall(0.5:0.95) \\ \hline\hline 
CircleNet~\cite{yang2020circlenet}   & 0.728   & 0.852   & 0.826   & 0.727  \\ 
CircleNet+  &\textbf{ 0.775 }  &\textbf{ 0.895 }  & \textbf{0.876}  & \textbf{0.776}  \\ 
 \hline
NMS~\cite{1699659}   & 0.641  & 0.776  & 0.730  & 0.636  \\ 
NMS+   & 0.719  & 0.828  & 0.803  & 0.717  \\ 
Soft-NMS~\cite{bodla2017soft}  & 0.570  & 0.661  & 0.635  & 0.565 \\ 
Soft-NMS+  & 0.616  & 0.699  & 0.686  & 0.613 \\ 
WCF(Ours)  & \textbf{\textcolor{lessblue}{0.823}} & \textbf{\textcolor{lessblue}{0.924}}  & \textbf{\textcolor{lessblue}{0.913}}  & \textbf{\textcolor{lessblue}{0.817}}  \\
WCF+ (Ours)  & \textbf{\textcolor{lessred}{0.873}} & \textbf{\textcolor{lessred}{0.951}}  & \textbf{\textcolor{lessred}{0.944}}  & \textbf{\textcolor{lessred}{0.873}}  \\
\bottomrule
\end{tabular}
\caption{Performance on rotation invariance: The chart displays the rotation invariance of various models and methods. From the results, we can see that the WCF method has achieved improvements in mean average precision and mean average recall. The results indicate that WCF possesses better rotation consistency.}
\label{table:rotation}
\end{table}

\subsection{Rotation consistency}
The study meticulously explores the rotation consistency of our object detection method, offering detailed insights in Table~\ref{table:rotation}. The results underscored the WCF method's notable consistency in rotation, highlighting its robustness against orientation changes. Our enhanced version of models also shows better rotation consistency compared to the original models.

\subsection{Manual Annotation vs. Human-in-the-loop Annotation}

To evaluate the efficiency of manual annotation compared to a human-in-the-loop approach, we conducted a time analysis for annotating 10 WSIs. The results demonstrates that the HITL method considerably improves annotation efficiency, requiring an average of 2.9 minutes per image compared to 9.23 minutes per image for manual annotation.


\section{Conclusion}
This work is the first to ensemble detection results for circle representation. We introduced a novel ensemble method, Weighted Circle Fusion (WCF), to refine predictions from multiple deep learning models. WCF demonstrated superior precision metrics, outperforming conventional benchmarks, especially in high-error contexts. Our findings highlight WCF's potential in reducing errors in circle representation, making it a valuable strategy for medical image analysis using optimized deep learning approaches.

\acknowledgments
This research was supported by NIH R01DK135597 (Huo), DoD HT9425-23-1-0003 (HCY), NIH NIDDK DK56942 (ABF). This work was also supported by Vanderbilt Seed Success Grant, Vanderbilt Discovery Grant, and VISE Seed Grant. This project was supported by The Leona M. and Harry B. Helmsley Charitable Trust grant G-1903-03793 and G-2103-05128. This research was also supported by NIH grants R01EB033385, R01DK132338, REB017230, R01MH125931, and NSF 2040462. We extend gratitude to NVIDIA for their support by means of the NVIDIA hardware grant. This work was also supported by NSF NAIRR Pilot Award NAIRR240055.

\bibliography{report} 
\bibliographystyle{spiebib} 

\end{document}